\documentclass[lettersize,journal]{IEEEtran}
\usepackage{amsmath,amsfonts}
\usepackage{algorithmic}
\usepackage{array}
\usepackage[caption=false,font=normalsize,labelfont=sf,textfont=sf]{subfig}
\usepackage{textcomp}
\usepackage{stfloats}
\usepackage{url}
\usepackage{verbatim}
\usepackage{graphicx,cite}
\usepackage{booktabs,amsthm}
\usepackage{tikz,graphicx}
\usepackage{balance}
\usetikzlibrary {calc,arrows.meta,graphs,shapes.misc,positioning}
\DeclareMathAlphabet{\mathbfcal}{OMS}{cmsy}{b}{n}

\def\BibTeX{{\rm B\kern-.05em{\sc i\kern-.025em b}\kern-.08em
    T\kern-.1667em\lower.7ex\hbox{E}\kern-.125emX}}

\begin{document}
\title{Computational Framework for Estimating Relative Gaussian Blur Kernels between Image Pairs}
\author{Akbar Saadat \thanks{The author works with the R\&D Department  of Iranian railways (RAI). This work has developed as an update to his academic research on ''Depth Finding by Image Analysis'' since 1995. \\E-mail: saadat\_a@rai.ir, Tel: (+98)9123840343}}

\maketitle
\markboth{Akbar Saadat, Under PAMI Review Process by  Submission ID a0514333-eab4-495c-983f-2582347a8d09.}{Main Manuscript}

\begin{abstract}
Over the past three decades, defocus has consistently provided groundbreaking depth information in scene images. However, accurately estimating depth from 2D images continues to be a persistent and fundamental challenge in the field of 3D recovery.  Heuristic approaches involve with the ill-posed problem for inferring the spatial variant defocusing blur, as the depth dependent blur cannot be distinguished from the inherent blur. Given a prior knowledge of the defocus model, the problem become well-posed with an analytic solution for the relative blur between two images, taken at the same viewpoint with different camera settings for the focus. The Gaussian model stands out as an optimal choice for real-time applications, due to its mathematical simplicity and computational efficiency. And theoretically, it is the only model can be applied at the same time to both the absolute blur caused by depth in a single image and the relative blur resulting from depth differences between two images. Following the earlier verification for Gaussian model  in \cite{ASaa2026}, this paper introduces a zero training forward computational framework for the model to realize it in real time applications. The framework is based on discrete calculation of the analytic expression of the defocused image from the sharper one for the application range of the standard deviation of the Gaussian kernels and selecting the best matches. The analytic expression yields multiple solutions at certain image points, but is filtered down to a single solution using similarity measures over neighboring points.The framework is structured to handle cases where two given images are partial blurred versions of each other. Experimental evaluations on real images demonstrate that the proposed framework achieves a mean absolute error (MAE) below $1.7\%$ in estimating synthetic blur values. Furthermore, the discrepancy between actual blurred image intensities and their corresponding estimates remains under $2\%$, obtained by applying the extracted defocus filters to less blurred images.

\end{abstract}


\section{Introduction}
\IEEEPARstart{C}{omputer} vision is the only solution for making an active interaction between a machine and its environment to control. It deals with two-dimensional images of a scene as input to extract the third dimension or depth at each image point as output. Since the initial introduction of focal gradients by A.P. Pentland \cite{APPe1987} as a novel source of depth information, the accurate estimation of depth from two-dimensional images has persisted as a significant challenge in the three-dimensional recovery domain. In the context of capturing images with a limited depth of field, the occurrence of defocus blur is inevitable. This is caused by the scene points being out of focus or shifted away from the camera's focal plane within the scene. The amount of shift is directly correlated with the depth of the scene points, according to the geometric optics. 

Throughout the past three decades, numerous methodologies have been proposed to address the DFD problem \cite{APPe1987, MSub1988,MSub1990,YXio1993, Jens1993,SKNa1994,MSub1994,ANRa1997,ANRa1998,MWat1998,ANRa1999,DRaj2003,ANRa 2004}. These researchers belong to a group that has made some of the most significant contributions to DFD techniques. They established the foundations for later advances and introduced multiple mathematical models and algorithms that have shaped the development of the field to feature definition and feature detection from heuristic features to  hardware modifications  \cite{ALev2007}, and to the key point descriptors \cite{MHas2019} such as speeded-up robust features (SURF) \cite{HBay2006}, pyramid histogram of oriented gradient (PHOG) \cite{Abos2007}, scale invariant feature transform (SIFT) (\cite{DGlo1999, DGLo2004}), and probabilistic graphs such as Conditional Random Field (CRF)\cite{JLaf2001} and Markov Random Field (MRF) \cite{GRCr1983}. These features were considered for depth estimation in a single image with parametric \cite{ASax2009}and non-parametric (\cite{CLiu2009,BLiu2010}) machine learning procedures. 

By the emergence of deep learning architectures all responsibilities for feature extraction, feature detection and  mapping features to depth delivered to the multi layers of Convolutional Neural Networks (CNN) \cite{FLiu2016,FLiu2015,DEig2014},  which  infer directly depth map from the image pixel values. In this approach, there is no basic difference between depth estimation and semantic labelling, as jointly performing both can benefit each other \cite{LLad2014}. CNNs have their own Limitations in 3D recovery of a scene image. With precisely calibrated architectures and hyper parameters, they can learn features from the training set from scratch during the training period. They cannot do anything more than extrapolation for what is beyond this limited space, even if given infinite time to completely learn the training set. The research in \cite{PWan2016} points out that CNNs often fail to ensure their predictions align with the planar regions depicted in the scene. Existing CNN architectures (e.g., VGG-16 \cite{KSim2015}) can not predict good surface orientations from depth, and pooling operations and large receptive fields makes current architectures perform poorly near object boundaries \cite{QiXL2020}. In order to succeed in challenging image regions, such as areas near depth discontinuities, thin objects and weakly textured zones, it is necessary to learn a broad range of principles and features that limit the possibility of focusing on important details \cite{WIqb2023}. To address the aforementioned limitations, these researchers devised a combination of conventional hand-crafted and deep learning-based methods, collectively referred to as hybrid techniques. In these methods, the predictions of deep networks are refined by the features extracted from the input image. Furthermore, the results of deterministic features, such as edges, in the sparse locations where the features are available, are replaced with the networks' predictions based on an image formation model.

Setting a model for image formation with a proven relationship to depth allows the reconstruction of a dense depth map entirely by a hand-crafted feature with no reliance on deep networks. The research conducted in \cite{ASaa2017} indicated that a hand-crafted feature with the Gaussian model for the defocusing operator could offer superior performance compared to learned features, including those derived from deep learning. The research conducted in \cite{ASaa2026} indicated that  “Defocus Aberration Theory Confirms Gaussian Model for Defocus Operator in Most Imaging Devices”. This research focused on the role of diffraction-limited optics in validating the Gaussian model for the defocus operator in conventional imaging systems,  revealed how diffraction-limited systems support the Gaussian approximation of the defocus operator.

The Gaussian model offers significant computational advantages and analytical tractability in both the spatial and frequency domains, rendering it particularly well-suited for real-time  DFD applications. From a theoretical standpoint, it is the unique formulation capable of jointly characterizing the absolute blur induced by scene depth in a single image and the relative blur arising from depth disparities between multiple images. Within conventional imaging systems, the Gaussian function serves as an accurate approximation of defocus operators, with deviations primarily attributable to the discrete sampling imposed by the image sensor grid. Although aliasing artifacts may arise, images satisfying the Nyquist sampling criterion, corresponding to the minimum spatial period of two pixels, remain within the reliable domain of application. Owing to its strong theoretical foundation, robustness to sampling constraints, and computational efficiency, the Gaussian filter constitutes an optimal choice for both software-based and hardware-accelerated implementations, thereby enabling large-scale image processing with real-time performance and informing the architectural design of dedicated fabrication modules.
 
This paper introduces a computational framework for estimating the standard deviation of the defocusing Gaussian filter that characterizes the transformation between an image point and its corresponding sharper reference. The organization of the paper is as follows. Section 2 examines the accuracy of the proposed framework on real images  by the effects of artificially applied defocus blur. Section 3 reviews fundamental aspects of the existing DFD theory with respect to image formation models. The formulation of the theory for discrete image pairs is presented in Section 4. Section 5 details the computational structure for estimating the Gaussian kernel associated with relative blur between two scene images. Experimental results validating the framework on real image data are reported in Section 6. Finally, Section 7 summarizes the key findings and concludes the paper.

\section{Framework's Accuracy}
The proposed computational framework is designed to extract the Gaussian depth-dependent relative blur between two images of a scene, referred to as the all-focused left image (L-Image) and the depth-dependent defocused right image (R-Image). The framework estimates the standard deviation of the Gaussian blur kernel at each image point, with the assumption that the standard deviation of the Gaussian kernel varies linearly from 1 to 2 across either hight or width of the image. To assess the framework’s accuracy, a structured evaluation using two real-world images is  employed , where the R-Image in each case is the result of artificially blurred the L-image using known standard deviation values. The extracted blur values are compared to ground truth blur parameters, with accuracy quantified through the MAE metric.
\begin{figure}[!t]
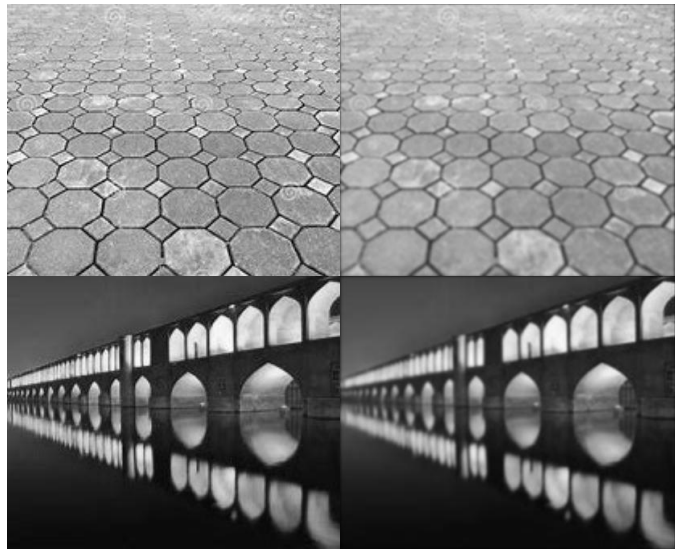

\centering
   \includegraphics[width=0.50\linewidth]{Fig1a\_Up\_ImgH}%
  \includegraphics[width=0.50\linewidth]{Fig1b\_Up\_ImgBH}\\
  \includegraphics[width=0.50\linewidth]{Fig1a\_Down\_ImgH}%
  \includegraphics[width=0.50\linewidth]{Fig1b\_Down\_ImgBH}
\caption{R-Images are the convolved results of L-Images using Gaussian blur kernels with a spatially-varying standard deviation increases linearly from 1 to 2. For the upper pair, this gradient is applied from top to bottom; for the lower pair, it is applied from left to right. The MAE for relative blur extraction or estimating the known standard deviation  is  $1.4\%$ for the upper image pair and $1.47\%$ for the lower pair.}
\label{Fig1}
\end{figure}

The upper images in Fig.\ref{Fig1} show a paved surface made up of interlocking tiles arranged in a repeating geometric pattern. The view is angled, creating a sense of depth as the pattern recedes into the distance. Considering the  L-Image in left as it  is entirely focused, the R-Image in right is an artificially blurred version of the L-Image. The standard deviation of the Gaussian blur  increases linearly from 1 at the topmost points to 2 at the downmost points, creating a structured progression in blur intensity. This setup provides a mathematically precise test case where the exact blur values are known at every image point, facilitating direct evaluation of extraction accuracy. The MAE  for relative blur computed for the images is $1.4\%$ demonstrating a high level of precision in extracting depth-dependent blur.

The lower images in Fig.\ref{Fig1} depicts Sio-Seh Pol, the iconic bridge in the city of Esfahan, Iran. Captured at night, the bridge’s series of illuminated arches stretch across the frame, casting striking reflections on the still surface of the water below. The repetition of the arches, mirrored almost perfectly in the river, creates a powerful sense of symmetry and rhythm. Let the L-Image is a sharp, focused capture of the scene, the R-Image is artificially blurred using the Gaussian standard deviation progression increasing linearly from 1 at the leftmost points to 2 at the rightmost points across the image width. This setup tests the framework’s robustness in extracting blur characteristics under realistic conditions. While maintaining strong performance, the extracted blur values exhibit slightly higher deviations from ground truth compared to the upper case, with a computed MAE of  $1.7\%$.

\section{DFD General Theory}
\noindent DFD theory is derived from the image formation model in geometric optics, which ignores the wave nature of light and treats it as rays. The imaging system in geometric optics is characterized by the parameters: $A$ for lens diameter or aperture, $f$ for focal length and $d_i$ for the distance of the image plane to the lens. In the model,  as shown in Fig.\ref{Fig1},  all rays parallel to optical axis converge to the focal point and all rays emerge from a single scene point on the focused plane illuminate the image point on the image plane. The fundamental equation of thin lenses describes the focusing distance $d_f$ in the scene by (\ref{LensLaw}).
\begin{equation}
\label{LensLaw}
d_f = \frac{f d_i}{d_i-f}
\end{equation}
\begin{figure}[!t]
\centering
\includegraphics[width=2.5in]{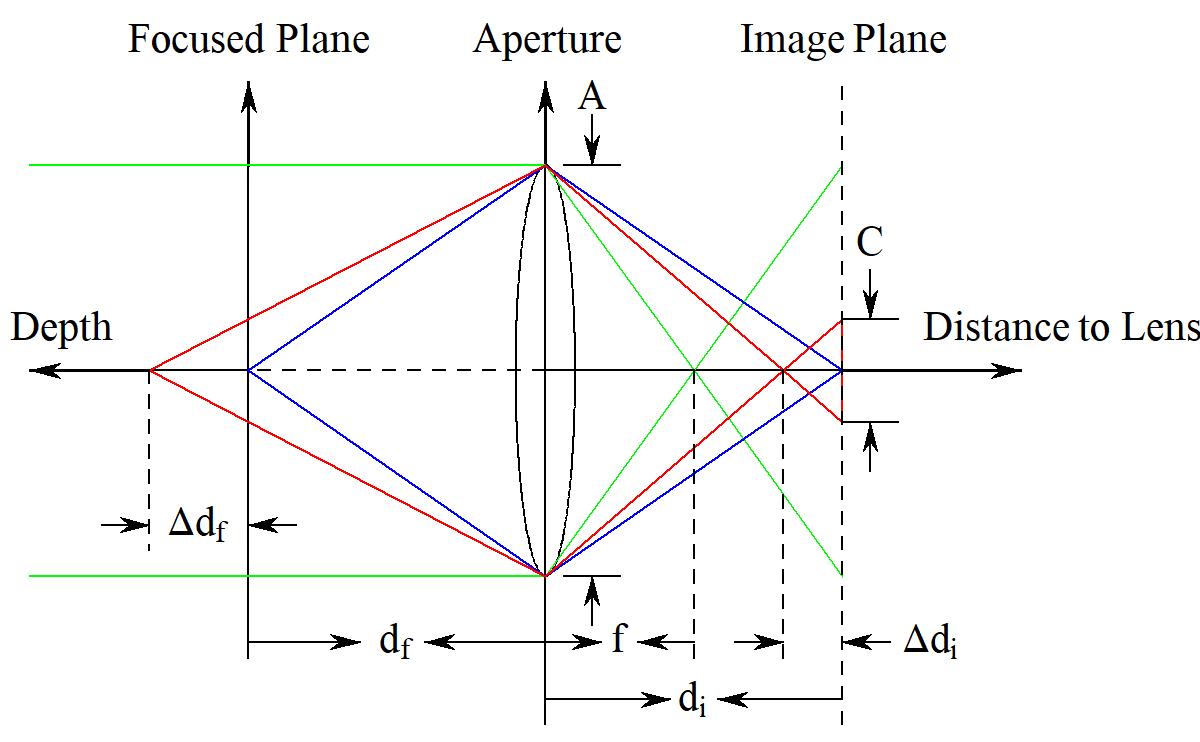}
\caption{Image formation through ray tracing in geometric optics. Both triples $(f,d_i,d_f)$ and $(f,d_i+\Delta d_i,d_f+\Delta d_f)$ are described by the lens law.}
\label{Fig2}
\end{figure}

Any displacement of a scene point from the focal plane by an amount $\Delta d_f$ leads to a corresponding shift of the focused image from the image plane by $\Delta d_i$. The updated position pair, $d_f+\Delta d_f$ for the scene point and $d_i+\Delta d_i$ for the focused image, remains governed by the thin lens law, as expressed in (\ref{Eq2}).
\begin{equation}
\label{Eq2}
d_f +\Delta d_f = \frac{f (d_i+\Delta d_i)}{d_i+\Delta d_i-f}
\end{equation}
This shift spreads the image point to the blur circle, designated as the Circle of Confusion (CoC) with a diameter C determined by similar triangles as (\ref{Eq3}).
\begin{equation}
\label{Eq3}
 \frac{C}{A} = \frac{\Delta d_i}{d_i+\Delta d_i}
\end{equation}
The signs of $\Delta d_i$ and $\Delta d_f$ are derived from the directions of the horizontal axes, designated as "Depth" and "Distance to Camera Lens," within rectangular coordinate systems with vertical axes positioned at the focal and aperture planes, as illustrated  in Fig.\ref{Fig2}.  Eliminating $\Delta d_i$    between  (\ref{Eq2}) and  (\ref{Eq3}) relates depth of the scene point $\Delta d_f$ to the blur circle $C$ through the settings parameters by (\ref{Eq4}).
\begin{equation}
\label{Eq4}
\Delta d_f  = \frac{C d_f}{C_o-C}  \hspace{1cm}  C_o=\frac{A f}{d_f-f} 
\end{equation}

The maximum value of $C$ at specified camera settings determines the admissible range validating the Gaussian model for depth estimation, with MAE as the error metric\cite{ASaa2026}. This extremum is dependent to the selected settings, as for the relative depths constrained within $-\eta < \Delta d_f < \eta$, direct manipulation of (\ref{Eq4}) leads to the derivation of (\ref{Eq5}).

\begin{align}
&C_{\text{max}} \stackrel{\triangle}{=}  \!m\!a\!x|C| = \label{Eq5} \\
&max \left|\frac{\Delta d_f}{d_f+\Delta d_f}\frac{Af}{d_f - f}\right|=\frac{\eta}{1-\eta} \frac{Af}{d_f - f}\nonumber
 \end{align}
Using $f_n=f /A$  reformulates (\ref{Eq5}) into a quadratic function of $f$ as (\ref{Eq6}).
\begin{equation}
\label{Eq6}
f^2+C_m f_n f -C_m f_n d_f = 0, \quad  C_m = \frac{1-\eta}{\eta}C_{\text{max}}
\end{equation} 
For the given $C_m$, this equation has two distinct real roots with opposite signs. The positive root is given by (\ref{Eq7}). 
\begin{equation}
\label{Eq7}
f = \frac{C_m f_n}{2}\left(\sqrt{1+\frac{4d_f}{C_m f_n}}-1\right)
\end{equation} 
The investigation into the camera settings for the relative depth $\eta=10\%$  in a practical ranges of $f_n$, $d_f$ and $C_{\text{max}}$   in \cite{ASaa2026} has led to MAE less than $1\%$ on conforming Gaussian model for defocus operator in the frequency domain. For any entity of  $(f_n, d_f, C_{\text{max}})$, the focal length given by $C_m$ in (\ref{Eq6}) and (\ref{Eq7}) completes the settings of the imaging device.

In Depth from Defocus applications, two particularly relevant focusing configurations are when the camera is focused either on the foreground $(d_f=d_F)$ or on the background $(d_f=d_B)$. When focused on the foreground, the depth $\Delta d_f$ ranges from $0$ at the foreground to $d_B-d_F$ at the background. For the  general case where the focal length $f$ is much smaller than $d_F$ (a first-order approximation), and further where $d_F$ is much smaller than $d_B$ (a second-order approximation), the maximum circle of confusion $C_{\text{max}}$ is given by Equation  (\ref{Eq8}).
\begin{equation}
\label{Eq8}
C_{\text{max}}= \frac{d_B-d_F}{d_B}\frac{Af}{d_F - f}\approx \frac{f^2(d_B-d_F)}{f_n d_B d_F }\approx \frac{f^2}{f_n d_F }
\end{equation}
 It can be verified that focusing on the background yields identical result as shown in (\ref{Eq8}). 
 
The relative size of $C_{\text{max}}$, expressed in terms of the image pixel pitch, is an important parameter governing the accuracy of the Gaussian kernel's standard deviation estimation. Excessively large values necessitate a larger computational image area, which adversely impacts processing speed. Conversely, values significantly smaller than one pixel pitch  fall below the discretization threshold and become undetectable by standard image-based calculations. Based on a trade-off between computational feasibility and speed, and supported by the demonstrated accuracy of the Gaussian model in our previous research, a relative size in the range of $(0.5, 5)$ is proposed for real-time applications. The high-resolution sensors available in modern imaging devices allow for image decimation, enabling us to rescale or decimate the input image in order to maintain the desired relative value for $C_{\text{max}}$.

\section{Formulating the Theory for\\ Discrete Image Pairs}
\noindent A foundational technique in DFD leverages a pair of images of a static scene captured from a fixed viewpoint but with differing focal settings. Let $i_L(x,y)$ represent the L-Image, serving as a reference captured at a baseline focal distance. The corresponding R-Image, denoted by $i_R(x,y)$, is acquired by adjusting the camera's focal plane to a different depth. The defocus process is mathematically modeled as the convolution of the L-Image with a depth-dependent point spread function (PSF), which is modeled by a Gaussian kernel $h(x',y';\sigma(x,y))$. The scale parameter $\sigma(x,y)$ of the kernel is a function of the spatial coordinates, directly linked to the local depth at each point. This relationship is formally expressed as 
\begin{align}
\label{Eq9}
&i_R(x,y) =  \iint\limits_{-\infty}^{+\infty}i_L(x-x',y-y')h(x',y';\sigma(x,y))dx'dy'\nonumber\\
&h(x',y';\sigma(x,y)) =  \frac{1}{2\pi\sigma^2(x,y)}\exp(-\frac{x'^2+y'^2}{2\sigma^2(x,y)})
\end{align}
The primary objective in DFD is to estimate the unknown depth map by inferring the spatially-varying parameter $\sigma(x,y)$ from the observed image pair $i_L(x,y)$ and $i_R(x,y)$. 
In order to exploit the inherent circular symmetry of the defocus kernel $h(x',y';\sigma(x,y))$ within the $(x',y')$ domain, it is preferable to reformulate the analysis in terms of radial coordinates. The kernel is consequently denoted by the radial function $H(r';\sigma(x,y))$, where $r' = \sqrt{x'^2 + y'^2}$. In accordance with this, the radial representations of the images are also defined, as specified in Equation (\ref{Eq10}).
\begin{align}
\label{Eq10}
&I(r) =  \frac{1}{2\pi}\int_{0}^{2\pi}i(r\cos\theta,r\sin\theta)d\theta\\
&(i;I) = (i_R,i_L;I_R,I_L) \nonumber
\end{align}
This transformation simplifies the subsequent integration operation required in Equation (\ref{Eq9}).
 
In the general case where the L-Image is not in focus, analytical DFD methods derive depth by solving a system of two equations formulated from a pair of images, $i_L (x,y)$ and $i_R (x,y)$, captured under different camera parameter settings. These images are modeled as the convolution of a latent sharp image with distinct defocus operators, $h_L (x',y';\sigma_L(x,y))$ and $h_R (x',y';\sigma_R(x,y))$, characterized by their spatially-varying blur parameters $\sigma_L (x,y)$ and $\sigma_R (x,y)$, respectively. The first equation in the pair is linear and is governed by the camera's settings, while the second is derived from the relationship between the relative blur observed in the two images. Collectively, these are termed the camera-based and image-based DFD equations, the solution of which yields the objective blur or depth at each image point.

This paper focuses on the relative blur $\sigma(x,y) = \sqrt{\sigma_R^2(x,y) - \sigma_L^2(x,y)}$ between the images, which constitutes a depth-dependent parameter in the general case. By shifting the origin of the coordinate system to the point of interest for relative blur estimation and denoting $\sigma(0,0)$ simply as $\sigma$, Equation~\ref{Eq9} establishes the relationship between the intensity value at a point in the R-Image and the corresponding points in the L-Image, as expressed in Equation~\ref{Eq11}.
\begin{equation}
\label{Eq11}
i_R(0,0) = \iint\limits_{-\infty}^{+\infty}i_L(x,y)h(x,y;\sigma)dxdy
\end{equation}
By utilizing the radial form of the L-Image derived in Equation~\eqref{Eq11}, the integral equation~\eqref{Eq12} is obtained for solving the relative blur parameter $\sigma$.
\begin{equation}
\label{Eq12}
I_R(0) = i_R(0,0) =\frac{1}{\sigma^2}\int\limits_{0}^{+\infty}I_L(r) r \exp(\frac{-r^2}{2\sigma^2})dr\
\end{equation}
To enable numerical computation, a practical integration limit must be established. The upper bound of integration is therefore set to $5\sigma$, which captures the dominant region of the blur kernel while maintaining computational efficiency. Under this constraint, $I_R(0)$ can be approximated by the function $\mathcal{M}(\sigma)$ as specified in Equation~\eqref{Eq13}.
\begin{equation}
\label{Eq13}
\mathcal{M}(\sigma) = \frac{1}{\sigma^2}\int_{0}^{5\sigma}I_L(r) r\exp(\frac{-r^2}{2\sigma^2}) dr
\end{equation}
The residual error introduced by this approximation is analytically bounded for the case where $I_L \leq 1$, with the bound given by Equation~\eqref{Eq14}.
\begin{align}
\label{Eq14}
I_R(0)-\mathcal{M}(\sigma) = &\frac{1}{\sigma^2}\int_{5\sigma}^{\infty} I_L(r) r \exp(\frac{-r^2}{2\sigma^2}) dr\leq  \nonumber \\ 
\frac{1}{2\sigma^2}\int_{5\sigma}^{\infty}r \exp(\frac{-r^2}{2\sigma^2})dr=&\frac{\exp(\frac{-25}{2})}{2}=1.86\times 10^{-6}
\end{align}
For practical implementation in discrete systems, the expression $\mathcal{M}(\sigma)$ is further simplified through a change of integration variable. This transformation yields the more computationally tractable form presented in Equation~\eqref{Eq15}.
\begin{equation}
\label{Eq15}
I_R(0) \approx \mathcal{M}(\sigma) = \int_{0}^{5} I_L(\sigma r) r \exp(\frac{-r^2}{2}) dr.
\end{equation}

\section{Computational structure for Estimating Relative Blur }
\noindent The integral in (\ref{Eq15}) is approximated using an $N$-point Riemann sum evaluated at $M$ uniformly spaced points $\sigma_m$, with $r_n$ denoting the corresponding evaluation points.
\begin{equation}
\label{Eq16}
\mathcal{M}(\sigma_m) \approx \sum_{n=1}^{N}  I_L(\sigma_m r_n) F(r_n), \hspace{3mm} m = 1,2,\hdots,M
\end{equation}
where $F(r)=5 r \exp(\frac{-r^2}{2})/N$. Let 
\begin{equation}
\mathbfcal{M} =
  \begin{bmatrix}
  \mathcal{M}(\sigma_1) \\
  \mathcal{M}(\sigma_2) \\
    \vdots \\
   \mathcal{M}(\sigma_M)
  \end{bmatrix},
\mathbf{I_L} = 
\begin{bmatrix}  I_L(\sigma_1r_1) \\
    I_L(\sigma_2 r_1) \\     \vdots \\     I_L(\sigma_M r_N)
  \end{bmatrix},
\mathbf{F} =
  \begin{bmatrix}
   F(r_1) \\
   F(r_2) \\
    \vdots \\
   F(r_N)
  \end{bmatrix}  \nonumber
\end{equation}
denote the vector representation of discrete values obtained from the corresponding functions. $\mathbf{I_L}$ is the vectorization of the $M \times N$ matrix $ \mathbfcal{I_L}$with the entities $\mathbfcal{I_L}(m,n)=I_L(\sigma_m r_n)$ denoted by $\mathbf{I_L}=\text{vec}(\mathbfcal{I_L})$.  The vectorization operation transforms a two-dimensional matrix into a column vector by stacking its columns sequentially. The inverse operation, which reconstructs a matrix from a vector by organizing elements into $M$ rows, is denoted as $\mathbfcal{I_L} = \mathrm{ivec}(\mathbf{I_L}, M)$. Using this notation, Equation~\eqref{Eq16} can be reformulated in matrix form as  in Equation~\eqref{Eq17}.
\begin{equation}
\label{Eq17}
 \mathbfcal{M}  =  \mathbfcal{I_L}\mathbf{F} = \text{ivec}(\mathbf{I_L},M)\mathbf{F}
\end{equation}

In accordance with (\ref{Eq9}),each element $I_L(\sigma_m r_n)$ in the matrix $\mathbfcal{I_L}$ represents the average intensity value of the local L-Image $i_L$ on the circle defined by $ r =\sqrt{x^2+y^2}= \sigma_m r_n$. 
This relationship imposes a constraint on the spatial support of the local L-Image, requiring a window size of $R_s \times R_s$ pixels. To ensure that the final matrix element $\mathbfcal{I_L}(M,N)$ can be properly computed, the window dimension must satisfy $R_s = 2\sigma_M r_N + 1$, which guarantees sufficient spatial coverage for the largest circle.

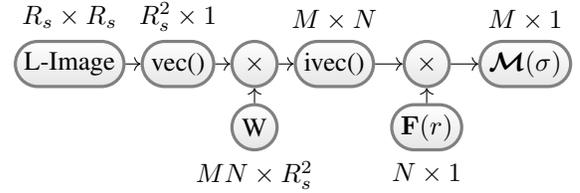
\begin{figure}
\centering
 \begin{tikzpicture}
   [node distance=5mm,   Block/.style={rectangle,minimum size=6mm,rounded corners=3mm,
                                       very thick, draw=black!50, top color=white,   bottom color=black!10}]
  \matrix[row sep=2mm,column sep=2mm] {
   \node (n11) [Block] {L-Image}; & \node (n12) [Block] {vec()};  & \node (n13) [Block] {$\times$};    
     & \node (n14) [Block] {ivec()};    &     \node (n15) [Block] {$\times$};     & \node (n16) [Block] {$\mathbfcal{M}(\sigma)$};  \\
    & &    \node (n23) [Block] { W};    &   &  \node (n25) [Block] {$\mathbf{F}(r)$}; & ;\\
      };                               
  \graph [grow right sep] {
    (n11) ->  (n12) ->  (n13) ->  (n14) -> (n15) ->  (n16) ;      (n23) ->  (n13);       (n25) ->  (n15) ; };
  \node (SizeLocImg) [above=1pt of  n11] {$R_s\times R_s$};    \node  (SizeVec) [above=0pt of n12] {$R_s^2\times 1$};
  \node  (SizeIV) [above=0pt of n14] {$M\times N$};   \node  (SizeBV) [above=0pt of n16] {$M\times 1$};
  \node  (SizeW) [below=0pt of n23] {$MN\times R_s^2$};  \node  (SizeFR) [below=1pt of n25] {$N\times 1$};
 \end{tikzpicture}
\caption{Computational framework for estimating a pixel's intensity in the R-Image. For a pixel centerd in a local $R_s \times R_s$ image patch in the L-Image, the image intensity in the R-Image for $M$ blur levels $\sigma_m$ is computed by approximating the underlying integral with using an $N$-point Riemann sum. Each block represents either a vector quantity or an operation, characterized by the specified dimensions of the original or transformed data as detailed in the main text.}
\label{Fig3}
\end{figure}

In the most general formulation, each element $I_L(\sigma_m r_n)$ can be expressed as a linear combination of all $R_s^2$ pixels within the local image window, as established in  (\ref{Eq18}).
\begin{equation}
\label{Eq18}
I_L(\sigma_m r_n) \approx \sum_{l=1}^{R_s^2} w_{mnl} i_{Ls}(l)
\end{equation}
The coefficient $w_{mnl}$ represents the weighting factor associated with parameters $\sigma_m$ and $r_n$ for the $l$-th element of the image, while $i_{Ls}(\cdot)$ denotes an arbitrary element from the sorted representation of $i_L(\cdot,\cdot)$. The product term in the argument of $I_L$ indicates that among the three indices of $w_{mnl}$, only two are independent, with the third being functionally dependent on the other two. 
When the sorting operation on  $ i_L(.,.)$  is realized by vectorization, the matrix presentation of (\ref{Eq18}) would be as (\ref{Eq19}).
\begin{equation}
\label{Eq19}
\text{vec}( \mathbfcal{I_L}) = {\bf{W}} \text{vec}(\mathbf{i_L}). 
 \end{equation}
$W$ is the $MN \times R_s^2$ weighting matrix composed of elements $w_{mnl}$. Applying the inverse vectorization operator $\mathrm{ivec}(\cdot, M)$ to both sides of (\ref{Eq19}) and substituting the resultant expression into (\ref{Eq17}) yields the computational framework presented in (\ref{Eq20}) and  illustrated graphically in Fig.\ref{Fig3}.
\begin{equation}
\label{Eq20}
 \mathbfcal{M}=\text{ivec}({\bf{W}}\text{vec}(\mathbf{i_L}),M)\mathbf{F},
 \end{equation}
This formulation provides a complete mathematical structure for the framework, pending explicit specification of the weighting elements $w_{mnl}$.

To determine the weighting elements $w_{mnl}$, the derivation begins by substituting (\ref{Eq18}) into (\ref{Eq16}). Interchanging the order of summations yields (\ref{Eq21}) .
\begin{equation}
\label{Eq21}
\mathcal{M}(\sigma_m) \approx  \sum_{l=1}^{R_s^2} i_{Ls}(l)  \sum_{n=1}^{N}  w_{mnl} F(r_n)  
\end{equation}
Since  $ w_{mnl}$ is independent of $n$ for fixed $m$ and $l$,  the inner summation reduces to (\ref{Eq22}),
\begin{align}
\label{Eq22}
&\sum_{n=1}^{N}w_{mnl}F(r_n) = w_{mnl} \sum_{n=1}^{N}F(r_n)  \approx \\ 
&w_{mnl} \int_{0}^{5} r \exp(\frac{-r^2}{2}) dr \approx w_{mnl}. \nonumber
\end{align}
Then (\ref{Eq21}) simplifies to  (\ref{Eq23}).
\begin{equation}
\label{Eq23}
\mathcal{M}(\sigma_m) \approx  \sum_{l=1}^{R_s^2} i_{Ls}(l)  w_{mnl}.  
\end{equation}
This expression represents the discrete realization of  (\ref{Eq11}) on the $R_s\times R_s$ grid. A direct comparison of these equations reveals that $w_{mnl}$ is given by  (\ref{Eq24}),
\begin{equation}
\label{Eq24}
w_{mnl}=\frac{H(r_l;\sigma_m)}{R_s^2}=\frac{\exp(-\frac{r_l^2}{2\sigma_m^2})}{2\pi R_s^2\sigma_m^2}
\end{equation}
thereby completing the definition of the weighting matrix $W$ within the proposed computational framework. 

\section{Experimental Results}
\noindent Building upon preliminary evaluations with synthetically defocused images, this section presents quantitative and qualitative assessments of the proposed framework's performance on real-world defocus imagery. The benchmark dataset employed in these experiments must satisfy several criteria essential for validating optical defocus characteristics: (1) It should contain paired images of identical scenes captured from the same viewpoint with varying focus distances; (2) All images must represent authentic photographic captures rather than synthetic renderings; (3) The blur manifestations must arise naturally from camera optics without artificial processing; (4) Each image pair should consist of a predominantly sharp L-Image  and a corresponding mostly defocused R-image.

\begin{figure}
\centering
 \begin{tikzpicture}
   [node distance=5mm,   Block/.style={rectangle,minimum size=6mm,rounded corners=3mm,
                                       very thick, draw=black!50, top color=white,   bottom color=black!10}]
  \matrix[row sep=2mm,column sep=2mm] {
   \node (n11) [Block] {Comp. Frame.}; & \node (n12) [Block] {$\mathbfcal{M}(\sigma)$};  & \node (n13) [Block] {Estimate $\sigma$};  & \node (n14) [Block] {R-Image}; \\
    \node (n21) [Block] {L-Image}; & & \node (n23) [Block] {Convolve};  & \node (n24) [Block] {$\hat{R}$-Image}; \\
      };                               
  \graph [grow right sep] {
    (n11) ->  (n12) ->  (n13) <-  (n14) ;      (n21) -> (n23) -> (n24);     (n21) ->  (n11) ; (n13) -> (n23) ; };
 \end{tikzpicture}
\caption{Performance evaluation of the computational framework can be done by comparing the actual R-Image  with its estimation $\hat{R}$-Image at the pixel level. The framework generates all pixel values as $\mathbfcal{M}(\sigma)$. The estimation  is produced by convolving the L-Image with the Gaussian kernel corresponding to the estimated $\sigma$ of that pixel.}
\label{Fig4}
\end{figure}
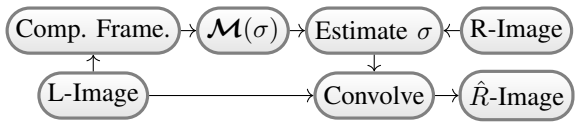

Given a dataset satisfying these requirements, the framework's performance is evaluated using the MAE between the actual R-image and its estimation. This estimation is obtained by applying the extracted defocus filters from the L-Image through the framework, as illustrated in Fig.~\ref{Fig4}. The procedure begins by processing the L-Image through the proposed computational framework (Comp. Frame.) shown in Fig.~\ref{Fig3}. This block generates blurred image values $\mathbfcal{M}(\sigma)$ corresponding to all candidate $\sigma$ values at each pixel. These values are subsequently compared with the R-Image pixel values to estimate the optimal $\sigma$ value from the candidates in $\mathbfcal{M}(\sigma)$. Finally, the estimated R-Image ($\hat{R}$-Image) is obtained by performing a convolution of the L-Image with the corresponding estimated Gaussian kernel at each pixel.

A review of prominent computer vision datasets for defocused imagery reveals that no single dataset satisfies all four established criteria. The datasets that come closest, fulfilling criteria 1--3, are the Dual-Pixel Defocus Deblurring (DPDD) dataset~\cite{AAbu2020} and the Real Multi-Focus Fusion (Real-MFF) dataset~\cite{JZha2020}. In contrast, the CUHK defocus dataset~\cite{LXuJ2010} satisfies criteria 1 and 2 but falls short on criterion 3, as its defocused images are synthetically generated by convolving all-in-focus images with depth-based kernels rather than capturing natural camera blur; it also does not satisfy criterion 4. Similarly, the Everything is Better with Bokeh Dataset (EBDB)~\cite{MABa2019} meets criteria 1 and 2, but its bokeh effects are often computationally applied, failing criterion 3, and it also does not meet criterion 4. The Real subset of the Bokeh-1k dataset satisfies criteria 1--3 akin to DPDD but fails criterion 4 due to the presence of in-focus regions within its blurry images. Finally, purely synthetic datasets such as SYNDOF~\cite{SSSK2017} do not satisfy criterion 2, as they are not composed of real-world imagery.
 
Real-MFF consists of realistic image pairs, where one image in each pair is focused on the background ($I_B$) and the other on the foreground ($I_F$). Each pair is accompanied by an all-in-focus image, which serves as the ground truth and is generated via image fusion.  Based on a sharpness measure computed at each image point, the subsets of points where $I_B$ is sharper than $I_F$ are identified as $BH$, and conversely, the subsets where $I_F$ is less sharp than $I_B$ are denoted as $BL$. Similarly, the sharper subsets of points of $I_F$ are labelled $FH$, and the less sharp subsets of points of $I_B$ are labelled $FL$. 

The process for estimating the error metrics $e_B$ and $e_F$ is illustrated in Fig.~\ref{Fig5}. It begins by detecting these sharpness-based subsets ($BH$, $BL$, $FH$, $FL$), which may comprise sporadic points rather than continuous regions. Subsequently, the less sharp subsets $\widehat{BL}$ and $\widehat{FL}$ are estimated using the disparity estimation process shown in Fig.~\ref{Fig4}, where $I_B$ and $I_F$ serve as the L-Image and R-Image for the respective subset. The process concludes by calculating the MAE between the original and estimated subsets to obtain the final error values $e_B$ and $e_F$. Image points with an identical sharpness measure in both $I_B$ and $I_F$ are not involved in estimating the relative blur between the two images and are hence excluded from the error calculations. The error calculation architecture in Fig.~\ref{Fig5} is general in application concept that supports any pair of images of a scene, without requiring that one be focused on the background and the other on the foreground. This includes, for example, the image pairs found in the DPDD dataset. The local sharpness at each pixel is evaluated by computing the standard deviation of the intensity values within a $3 \times 3$ neighborhood centerd on that pixel.

Experimental results on sample image pairs from the Real-MFF and DPDD datasets are presented in Fig.~\ref{Fig6} and Fig.~\ref{Fig7}, respectively. In each figure, the top row displays the original image pair ($I_F$ and $I_B$) from the dataset. The bottom left image shows the exclusive diffusion regions of the original less-focused images: either $FL$ (where $I_F$ is less sharp than $I_B$) or $BL$ (where $I_B$ is less sharp than $I_F$). Similarly, the bottom right image shows the exclusive diffusion of the estimated less-focused images ($\hat{FL}$ or $\hat{BL}$) in the corresponding regions. The complementary regions in both bottom images--where $I_F$ and $I_B$ exhibit equivalent sharpness measures are replaced by the pixel-wise average of the original image pair.

The image shown in Fig.~\ref{Fig6}, captured with a Lytro Illum camera from the REAL-MFF dataset, conforms to the imaging system specifications reported in~\cite{Lytr2015}. The sensor has dimensions of $10.82 \,\text{mm} \times 7.52 \,\text{mm}$ and is covered by a micro-lens array of $625 \times 433$ elements, each recording the light field with an angular resolution of $15 \times 15$ ray directions.
These 4-D rays are mapped to a $2450 \times 1634$ pixel image on the sensor at a specific focus distance, corresponding to an approximate pixel pitch of $P = 4.5~\mu\text{m}$. The lens system features an 8× optical zoom with a 35-mm equivalent focal length range of $(30–250)$ mm, translating to an actual focal length range of $(9.5–77.8)$ mm, while maintaining a constant aperture of $f_n=2$. The image under analysis has dimensions of $433 \times 635$ pixels, with a foreground distance of approximately $d_F = 1000~\text{mm}$. 
By substituting these parameters into Eq.~\eqref{Eq8} at the harmonic mean focal length of $f = 17\,\text{mm}$, the maximum circle of confusion is obtained as $C_{\text{max}} = 32$ pixel pitches. This value must be scaled to below $5$ to conform to the proposed range for operating the defocusing operator under a Gaussian model.

Figure~\ref{Fig7} shows an image from the DPDD dataset, captured with a Canon EOS 5D Mark IV camera at one-fourth of its native resolution. The camera employs a full-frame CMOS sensor measuring $36 \,\text{mm} \times 24 \,\text{mm}$, with a native resolution of $6720 \times 4480$ pixels, each incorporating dual photodiodes for phase-detection autofocus. The downsampled resolution of $1680 \times 1120$ pixels corresponds to an approximate pixel pitch of $P = 21.43\,\mu\text{m}$. The aperture is set to $f_n=4$ for standard images and $f_n=22$ for the all-in-focus counterpart in each pair. Although the specific lens is not reported, the focal length varies across pairs and is consistent with a standard to short telephoto lens in the $(50,100)$ mm range. The nearest in-focus object, a tree trunk, is located approximately $1.0–1.2$ meters from the camera, giving an estimated foreground distance of $d_F = 1100~\text{mm}$. Substituting these values into Eq.~\eqref{Eq8} at the harmonic mean focal length of $f=67$ mm results in a maximum circle of confusion of $C_{\text{max}} = 48$ pixels.This value must be scaled to fewer than $5$  to meet the constraints for approximating the defocus operator with a Gaussian model.

Decimation is the fundamental operation in high resolution digital images for adopting $C_{\text{max}}$ to processing requirement, where the spatial resolution of an image is reduced by resampling at a lower rate. Prior to resampling, it is essential to apply a low-pass filter in order to suppress high-frequency components that would otherwise cause aliasing artifacts. The ideal low-pass filter, represented by the sinc function in the spatial domain, attenuates frequencies beyond the Nyquist limit, thereby ensuring that the downsampled image preserves only valid spectral content. However, this filter is of Infinite Impulse Response (IIR) type. Consequently, for a decimation factor $D$, a linear-phase digital filter is implemented using a Finite Impulse Response (FIR) Kaiser-windowed sinc function, as expressed in Eq.~\eqref{Eq25}
\begin{equation}
\label{Eq25}
g(k) = \text{sinc}\left(\frac{k}{D}\right)  \text{Kaiser}(2L+1, \beta), \hspace{3mm}-L\leq k\leq L
\end{equation}
 with parameters $L = 8$ and $\beta = 10$. The window function is characterized by a relative sidelobe attenuation of $-47.1\,\mathrm{dB}$ and a mainlobe width of $0.22$ at $-3\,\mathrm{dB}$, which provides an effective compromise between frequency selectivity and computational efficiency. By applying this filter before decimation, aliasing is minimized and the integrity of the image content is preserved in the reduced-resolution representation.

A summary of the key experimental statistics is provided in Table~\ref{tab:stats}, with detailed analyses for two specific cases and two consecutive decimation processes using a factor of 2 for each case. The pixel distribution, categorized based on whether the background $I_B$ is sharper than, equally sharp as, or less sharp than the foreground $I_F$, is approximately $(50\%, 23\%, 27\%)$ for Fig.~\ref{Fig6} and $(76\%, 0\%, 24\%)$ for Fig.~\ref{Fig7}.  Accordingly, for Fig.~\ref{Fig6} and Fig.~\ref{Fig7}, $(50\%, 76\%)$ of the pixels contribute to the error metric $e_{F}$, $(27\%, 24\%)$ contribute to $e_{B}$, and the remaining $(23\%, 0\%)$ do not contribute to either metric.
The maximum absolute errors for $(e_B, e_F)$ are $(0.021, 0.025)$ for Fig.~\ref{Fig6} and $(0.023, 0.014)$ for Fig.~\ref{Fig7}. The monotonically decreasing errors for the REAL-MFF database with increasing decimation factor or decreasing resolution indicate that $157 \times 109$ is a proper resolution that balances error. Similarly, tracing the monotonicity in the results for the DPDD dataset reveals $210 \times 140$ as a suitable candidate for resolution.

\begin{table}[!ht]
    \centering
    \caption{The experimental results on an image sample of two datasets.}
    \label{tab:stats}
    \begin{tabular}{lccccc}
        Dataset & Image & $I_B$ Sharper & $I_F$ Sharper& $e_B$ &  $e_F$ \\
                 & Resolution &than $I_F$&than $I_B$ &   &  \\ \toprule
        Real MFF &$625\times433$& $50\%$  & $27\%$ & $0.021$& $0.025$ \\ 
        Real MFF &$313\times217$& $43\%$ & $39\%$ & $0.016$& $0.012$ \\ 
        Real MFF &$157\times109$& $44\%$ & $46\%$ & $0.014$& $0.009$ \\ 
        DPDD     &$420\times280$& $76\%$ & $24\%$ & $0.023$& $0.014$ \\ 
        DPDD    &$210\times140$& $67\%$ & $33\%$ & $0.018$& $0.010$ \\ 
        DPDD    &$105\times70$& $58\%$ & $42\%$ & $0.017$& $0.016$ \\ 
    \end{tabular}
\end{table}

\section{Discussion and Conclusions}
\noindent  It is demonstrated in \cite{ASaa2026} that for the typical set of focused depths between  $1$ and $100$ meters, with  maximum depth variation $10\%$ of the focused depth, the defocus operators can be approximated by  Gaussian model with  less than $1\%$  MAE over the frequency domain up to one cycle per pixel pitch in most imaging devices. The Gaussian model can be applied theoretically  to both the absolute blur caused by depth in a single image and the relative blur resulting from depth differences between two images.  This paper presented the computational framework for estimating the standard deviation of the defocusing Gaussian filter at any image point from a sharper version of that. 

The computational framework in Fig.\ref{Fig3} calculates an approximation of the analytic expression of the  defocused image $\mathcal{M}(\sigma_m)$ in  (\ref{Eq16}) from the sharper image in the application range of the standard deviation of the Gaussian kernels and enables one to select the best match to the defocused image at each point. The practical processing dimensions of the framework for $N=100$,  $r_N=5$, $M=50$, $\sigma_1=0.1$ and $\sigma_M=5$ in the experiments was operating on the $51\times51$ local images with the $5000\times2601$ weighting matrix $W$ in the first layer and with  the $100\times1$ weithing matrix $F$ in the second layer, with required data reshaping to adopt vector productions.

\begin{figure}
\centering
\begin{tikzpicture}
\def\MinW{1cm}   \def\MinH{1.2cm}
\tikzset{
     >={Stealth[round]}, black!50, text=black, thick,
    every new->/.style  = {shorten >=1pt},
    front path/.style={  to path={-- ++(4mm,0)--++(0,#1)-- (\tikztotarget) \tikztonodes}},
    back path/.style={  to path={-- ++(-4mm,0)--++(0,#1)-- (\tikztotarget) \tikztonodes}},
    block/.style = {rectangle,align=center, minimum width=\MinW,  minimum height=\MinH}, 
    drawme/.style = {draw=black!50,  very thick,top color=white, bottom color=black!20,rounded corners},       
    Del/.style = {circle, drawme, minimum size=6mm}}
\newcommand{\MakeAnchors}[1]{%
  \coordinate (#1-wdon) at ($ (#1.south west)!0.5!(#1.west) $);
  \coordinate (#1-wup) at ($ (#1.west)!0.5!(#1.north west) $);
  \coordinate (#1-wtop) at ($ (#1.north west)!0.5!(#1.north) $);
  \coordinate (#1-etop) at ($ (#1.north)!0.5!(#1.north east) $);
  \coordinate (#1-eup) at ($ (#1.north east)!0.5!(#1.east) $);
  \coordinate (#1-edon) at ($ (#1.east)!0.5!(#1.south east) $);
}

\matrix[row sep=1.5mm,column sep=2mm] (m) {
\node (A) [block,drawme] {Estimate\\ BL}; & &\node (B) [block,drawme]  {Detect\\ Sharper}; & &\node  (C) [block,drawme]  {Estimate\\ FL};\\
& \node (D) [Del] {.};  & & \node (E) [Del] {.}; \\
};

 \node (eB) [circle, minimum size=2mm] at ($(D.south)+(0,-0.6)$) {$e_B$};
 \node (pB) [circle,font=\footnotesize] at ($(D.north)+(-0.2,0.1)$) {+};
 \node (nB) [circle,font=\footnotesize] at ($(D.west)+(-0.1,0.15)$) {-};
 \node (eF) [circle, minimum size=2mm] at ($(E.south)+(0,-0.6)$) {$e_F$}; 
 \node (pF) [circle,font=\footnotesize] at ($(E.north)+(0.2,0.1)$) {+};
 \node (nF) [circle,font=\footnotesize] at ($(E.east)+(0.1,0.15)$) {-};
 
\path (A); \pgfgetlastxy{\xA}{\yA};    \path (D); \pgfgetlastxy{\xD}{\yD}; 

 \node (BLhat) [circle, minimum size=2mm] at ($(A)+(0,\yD-\yA-9)$) {$\widehat{BL}$};
 \node (FLhat) [circle, minimum size=2mm] at ($(C)+(0,\yD-\yA-9)$) {$\widehat{FL}$};
       
\MakeAnchors{A}  \MakeAnchors{B}   \MakeAnchors{C}

 \node (IB) at ($(B-wtop)+(0,0.65)$) {$I_B$};
 \node (IF) at ($(B-etop)+(0,0.65)$) {$I_F$};  

\draw[->]  (B-wup) -- node[midway, above] (BHm) {BH} (A-eup);
\draw[->]  (B-wdon) -- node[midway, above] (BLm) {BL} (A-edon);
\draw[->]  (B-eup) -- node[midway, above] (FHm) {FH} (C-wup);
\draw[->]  (B-edon) -- node[midway, above] (FLm) {FL} (C-wdon);
\draw[->] (D.south)--++(0,-0.4);   \draw[->] (E.south)--++(0,-0.4);

\graph {
(BLm) ->  (D); (FLm) -> (E);(C.east) ->[ front path=\yD-\yA] (E.east);
(A.west) ->[back path=\yD-\yA] (D.west);(IF) ->(B-etop);   (IB) ->(B-wtop) ; };
   
\end{tikzpicture}
\caption{Error calculation architecture for image pairs ($I_B$, $I_F$). For the detected sharper subsets (BH and FH), their respective images ($I_B$ and $I_F$) serve as the L-image. The complementary, less sharp subsets (BL and FL) are then estimated as $\widehat{BL}$ and $\widehat{FL}$ using the process in Fig.~\ref{Fig4}. The final errors $e_B$ and $e_F$ are obtained by calculating the difference between the original and estimated less sharp subsets.}
 \label{Fig5}
\end{figure}
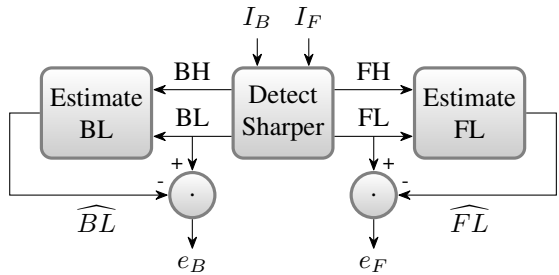

\begin{figure}[!t]
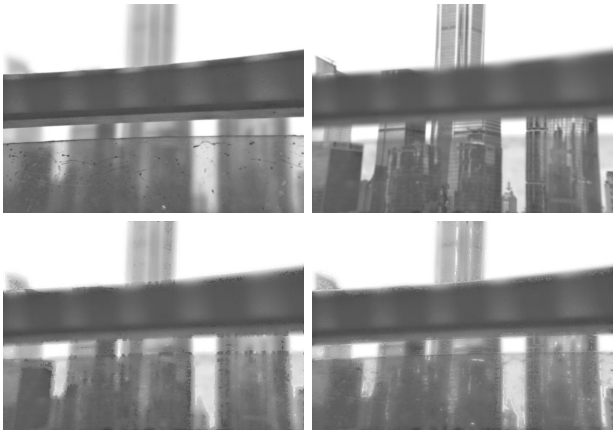

\centering
 \begin{tabular}{@{}c@{}c}
   \includegraphics[width=0.45\linewidth]{Fig6a\_1\_ImgA} %
  \includegraphics[width=0.45\linewidth]{Fig6a\_2\_ImgB}\\
  \includegraphics[width=0.45\linewidth]{Fig6a\_3\_Ib} %
  \includegraphics[width=0.45\linewidth]{Fig6b\_3\_Ibhat}
\end{tabular}
\caption{Analysis of a multi-focus image from the REAL-MFF dataset~\cite{JZha2020}, acquired with a Lytro Illum camera. Top row: foreground-focused (left) and background-focused (right) images. Bottom row: fusion of defocused regions (left) and its re-estimation via blur detection (right), as described in Fig.~\ref{Fig4}.}
 \label{Fig6}
\end{figure}

\begin{figure}[!t]
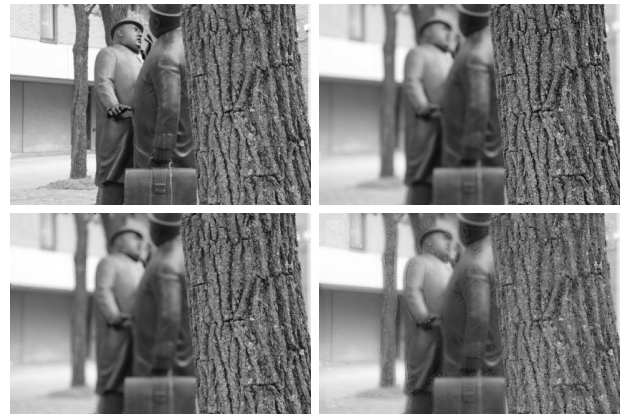

\centering
 \begin{tabular}{@{}c@{}c}
   \includegraphics[width=0.45\linewidth]{Fig7a\_1\_ImgA} %
  \includegraphics[width=0.45\linewidth]{Fig7b\_2\_ImgB}\\
  \includegraphics[width=0.45\linewidth]{Fig7a\_3\_Ib} %
  \includegraphics[width=0.45\linewidth]{Fig7b\_3\_Ibhat}
\end{tabular}
\caption{Experiment on an  image from the DPDD dataset~\cite{AAbu2020}, captured with a Canon EOS 5D Mark IV. Top row: all-in-focus at $f_n=22$ (left) and standard at $f_n=4$ (right) images. Bottom row: fusion of defocused regions (left) and its re-estimation using detected blur values (right), as detail in Fig.~\ref{Fig4}.}
\label{Fig7}
\end{figure}

The method for evaluating the performance of the framework in Fig.~\ref{Fig4} is based on estimating the defocused image point at each image point given the focused image at all points. The estimated R-Image ($\hat{R}$-Image) is obtained by performing a convolution of the L-Image with the corresponding estimated Gaussian kernel at each pixel. The experiment generalised the space of the image pairs from one image is less focused than the other at all points  to at any subset of points. The proposed architecture for error calculation in Fig.~\ref{Fig5} is on the bases this generalisation and supports any pair of images of a scene. Partitioning an image pair into background focused and foreground focused in the dataset Real MFF is an examples of this generalisation.
 
The approximation  for Gaussian model was studied under diffraction-limited conditions with defocus aberration. All lens aberrations, including those associated with monochromatic light (spherical aberration, coma, astigmatism, field curvature, image distortion) and chromatic aberrations (dispersion), and image noise (thermal, impulsive, shot, quantization) were not considered in the model.  
The computational framework is verified by real images in two cases. In the first case with known synthetic blur, the blur detection was realized with less than $1.7\%$ error and  re-estimating the defocused image with less than $0.0002$ error for the unity maximum value. In the second case with depth dependent unknown blur, retrieving the original defocused image points by the estimated blur was done with less than $0.02$ error. Although both values are small in absolute terms, the transition from synthetic to realistic data corresponds to a hundredfold increase in error. Nevertheless, the absolute magnitude of the error in the realistic case remains low, making it reasonable to highlight the high performance level of the proposed computational framework in mitigating lens aberrations and image noise under real-world conditions.

Theoretical supports in \cite{ASaa2026}, large size of local images and independent pixel blur assumption  for the computational framework makes each run  for an image pixel similar to an independent run for the framework. Therefore, the experiment on all other images of the datasets will not faced with noticeable error level deviations from what is experienced. The proposed architecture for error calculation in  Fig.\ref{Fig5} plays a key role for supporting both this argument and the performance stability of the computational framework over any image pairs. Based on the fabrication module for realizing the framework in  Fig.\ref{Fig3}, the research is ongoing on detecting the scene points in a predefined depth range from the related focused image pairs at the range end points.

\begin{IEEEbiography}[{\includegraphics[width=1in,height=1.25in,clip,keepaspectratio]{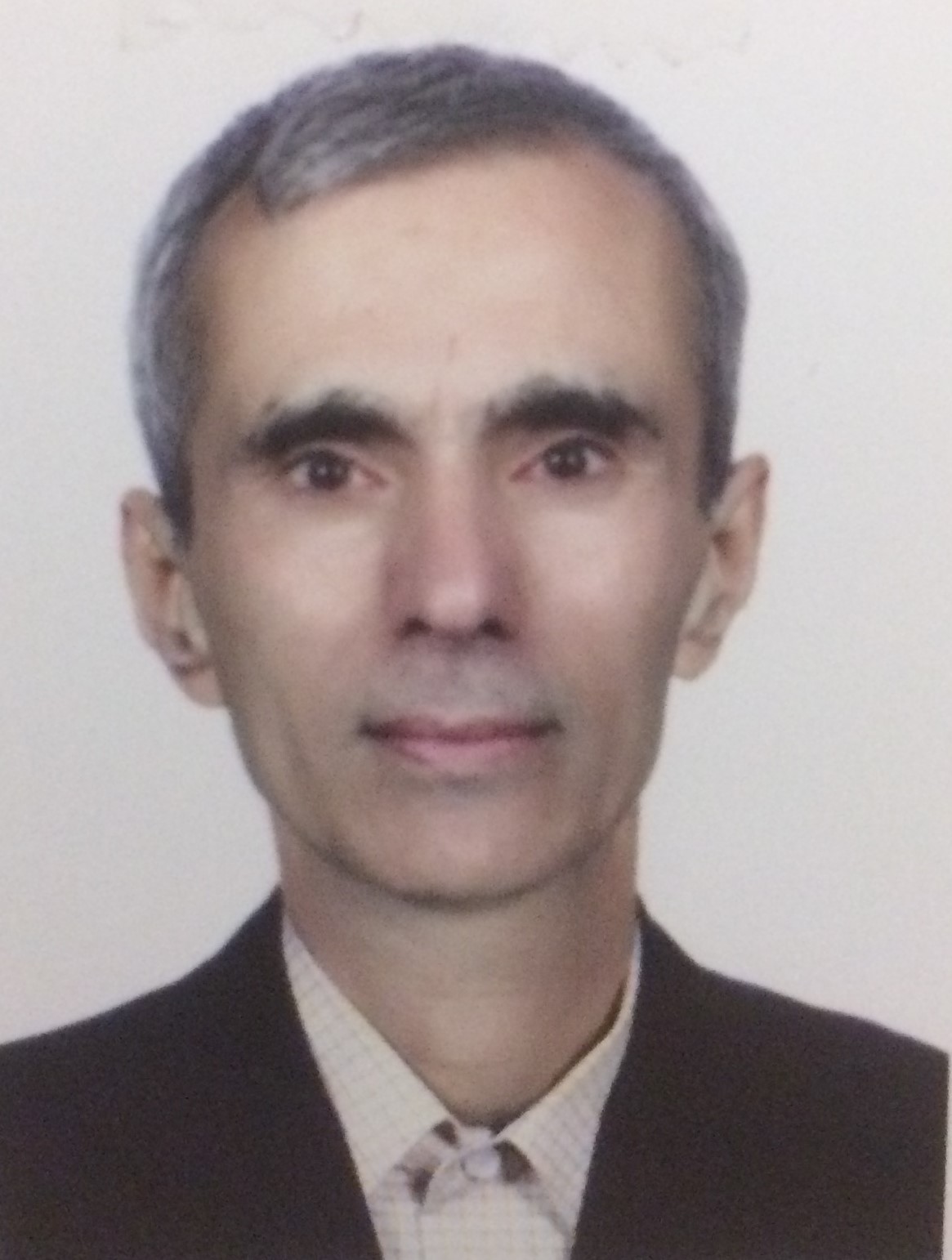}}]
Akbar Saadat  received the B.S. and M.S. degrees from Esfahan University of Technology, in 1986 and 1991, and the Ph.D. degree from Sharif University of Technology, in 1997, all in electrical engineering, Iran. He has been faculty member of Electrical Engineering Department in Esfahan University of Technology and Yazd University. Since 1998, he has been with the R\&D, Signalling, and Infrastructure Technical Departments at Iranian Railways. He is an expert in railway signalling and has completed several international training courses in Germany, India, and Japan. His research interests include computer vision, image analysis, and information processing.
\end{IEEEbiography}

\end{document}